\documentclass[11pt,a4paper]{article}
\usepackage[hyperref]{acl2020}
\usepackage{times}
\usepackage{latexsym}

\usepackage[utf8]{inputenc} 
\usepackage[T1]{fontenc}    
\usepackage{url}            
\usepackage{booktabs}       
\usepackage{amsfonts}       
\usepackage{nicefrac}       
\usepackage{microtype}      
\usepackage{enumitem}
\usepackage{color}
\usepackage{graphicx}
\usepackage{mathtools}
\usepackage{xcolor}
\usepackage{xspace}

\usepackage{microtype}

\aclfinalcopy 

\renewenvironment{quote}{%
   \list{}{%
     \leftmargin0.1cm   
     \rightmargin\leftmargin
   }
   \item\relax
}
{\endlist}

\newcommand{\suppai}{\texttt{SUPP.AI}\xspace}

\title{\suppai: Finding Evidence for Supplement-Drug Interactions}
  
\author{
  Lucy Lu Wang, Oyvind Tafjord, Arman Cohan, Sarthak Jain, Sam Skjonsberg, \\ \textbf{Carissa Schoenick, Nick Botner, Waleed Ammar} \\
  Allen Institute for AI \\
  Seattle, WA 98103 \\
  \small{\texttt{\{lucyw, oyvindt, armanc, sarthakj, sams, carissas, nickb, waleeda\}@allenai.org}}
}

\date{}

\begin{document}
\maketitle
\begin{abstract}
Dietary supplements are used by a large portion of the population, but information on their pharmacologic interactions is incomplete. To address this challenge, we present \suppai, an application for browsing evidence of supplement-drug interactions (SDIs) extracted from the biomedical literature. We train a model to automatically extract supplement information and identify such interactions from the scientific literature. To address the lack of labeled data for SDI identification, we use labels of the closely related task of identifying drug-drug interactions (DDIs) for supervision. We fine-tune the contextualized word representations of the RoBERTa language model using labeled DDI data, and apply the fine-tuned model to identify supplement interactions. We extract 195k evidence sentences from 22M articles (P=0.82, R=0.58, F1=0.68) for 60k interactions. We create the \suppai application for users to search evidence sentences extracted by our model. \suppai is an attempt to close the information gap on dietary supplements by making up-to-date evidence on SDIs more discoverable for researchers, clinicians, and consumers.
\end{abstract}

\section{Introduction}

More than half of US adults use dietary supplements \citep{kantor_trends_2016}. Supplements include vitamins, minerals, enzymes, and other herbal and animal products. Supplements and pharmaceutical drugs, when taken together, can cause adverse interactions \citep{sprouse_pharmacokinetic_2016, asher_common_2017, ronis_adverse_2018}. Some studies describe the prevalence of supplement-drug interactions (SDIs) in the hospital setting \citep{levy_interactions_2016, levy_adverse_2017, levy_potential_2017} or among groups such as patients with cancer \citep{alsanad_cancer_2014}, cardiac disease \citep{karny-rahkovich_dietary_2015}, HIV/AIDS \citep{jalloh_dietary_2017}, or Alzheimer's disease \citep{spence_brief_2017}. However, these studies largely rely on manual curation of the literature, and are slow and expensive to produce and update. It is also difficult to aggregate their results, and researchers, clinicians, and consumers can lack appropriate up-to-date information to make informed decisions about supplement use.

A resource that provides experimental evidence for SDIs could serve as a good intermediary tool, allowing experts to quickly access information and translate it for healthcare providers and consumers. Such a tool could ease the bottleneck of manual curation by directing researcher attention to the most pertinent and novel interactions appearing in recent trials and case reports. Our goal is to create such a resource using state-of-the-art methods in NLP and IE, and allow users to better identify appropriate uses of supplements as well as risks for SDIs.

Automated approaches have been used to extract drug-drug interactions (DDIs) from literature and other documents \citep{Tari2010DiscoveringDI, Percha2011DiscoveryAE, segura-bedmar_linguistic_2011, Kim2014ExtractingDI, Zhang2016LeveragingSA, noor_drug-drug_2017, Lim2018DrugDI}, complementing broadly-used but primarily manual methods \cite{grizzle_identifying_2019}. We expand upon this work to automatically extract evidence for SDIs, as well as supplement-supplement interactions (SSIs), from a large corpus of 22M biomedical and clinical texts derived from Semantic Scholar.\footnote{\href{https://www.semanticscholar.org/}{https://www.semanticscholar.org/}} We leverage labeled datasets for DDI identification for supervision, and train a model that transfers to the related task of identifying supplement interactions. We surface the resulting evidence on \suppai for browsing and search.

To summarize, our contributions are: \vspace{-1mm}
\begin{enumerate}[itemsep=0.1mm]
    \item A model for identifying SDI/SSI evidence
    \item A dataset of 195k evidence sentences supporting supplement interactions, publicly accessible for download or via a web API, and
    \item \suppai, an application for browsing and searching the extracted evidence.
\end{enumerate}

\begin{figure*}[tbph!]
    \centering
    \includegraphics[width=0.45\textwidth]{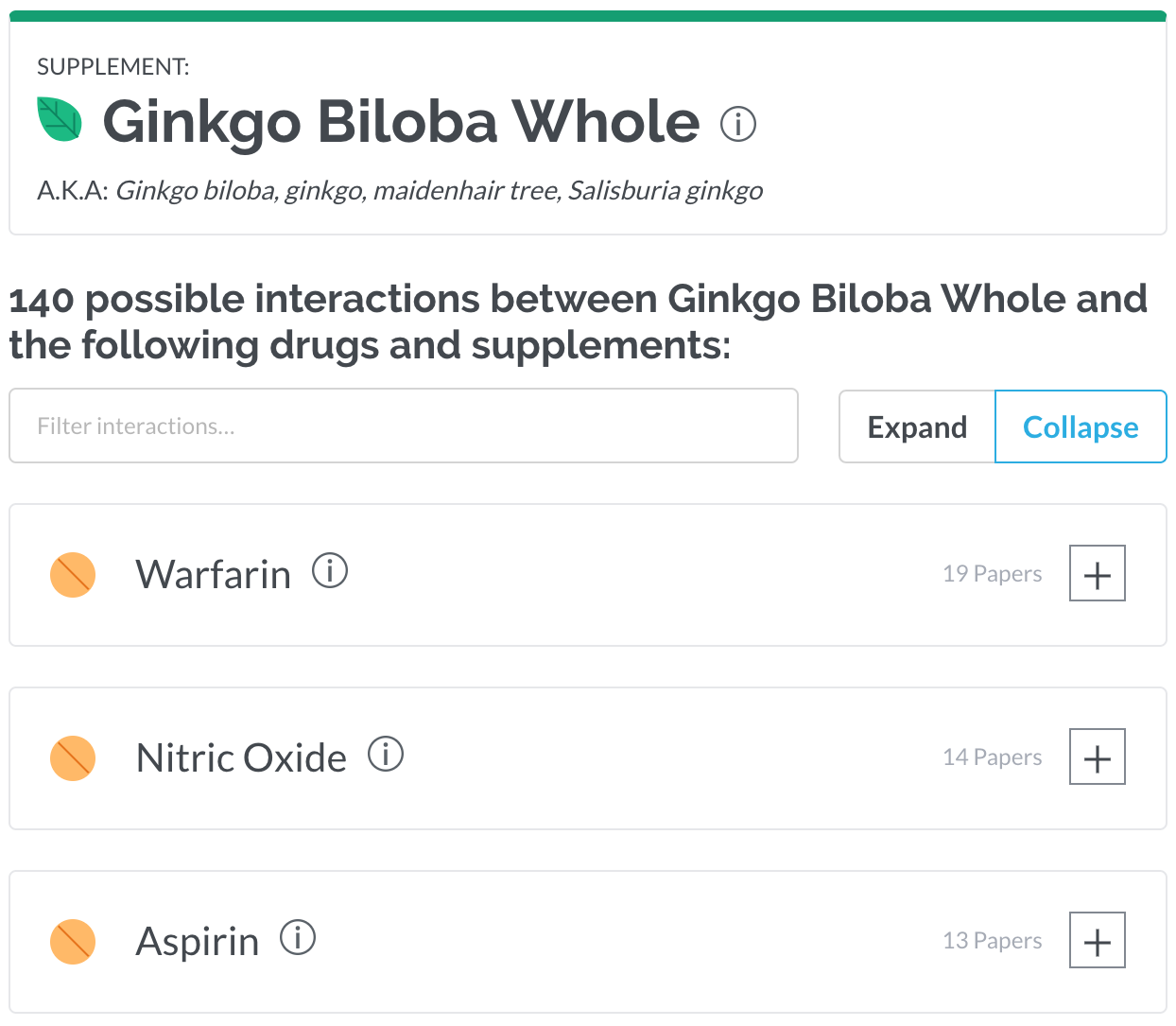}\hspace{2mm}\includegraphics[width=0.52\textwidth]{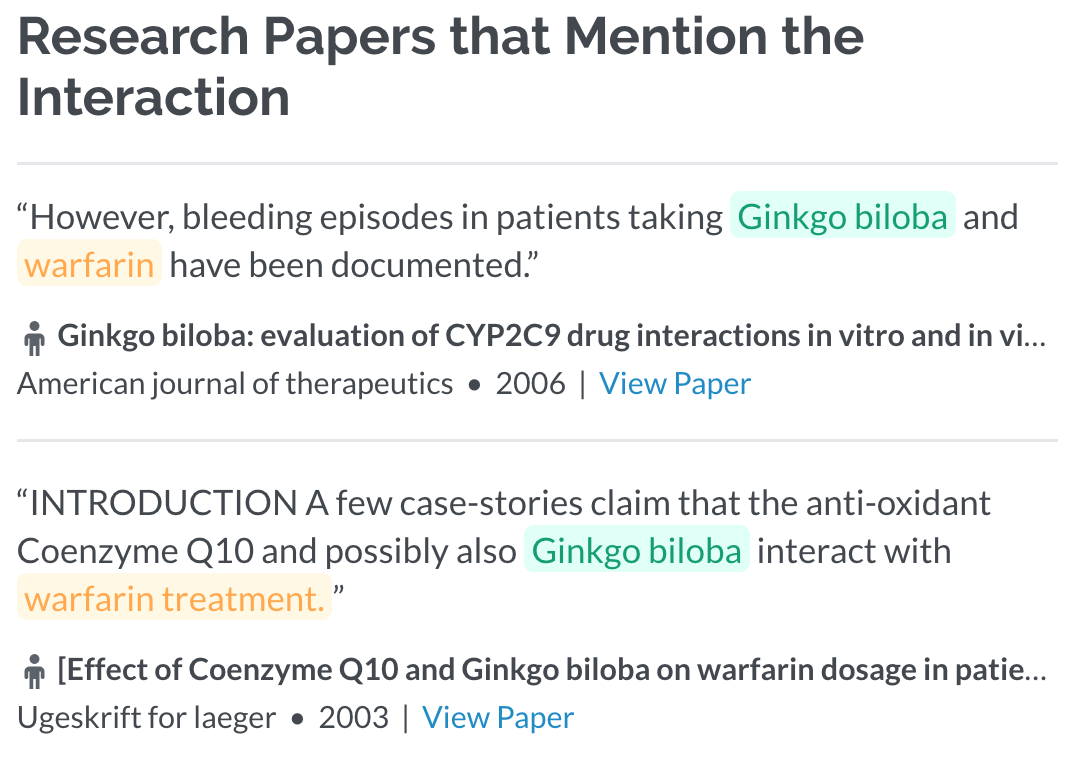}
    \caption{Top results for interactions with Ginkgo (\emph{left}), and top evidence sentences for the SDI between Ginkgo and Warfarin (\emph{right}). Source paper metadata are given below each evidence sentence.}
    \label{fig:supp_ai}
\end{figure*}

\section{Supplement interaction browser}
\label{sec:supp_ai_browser}
Information on supplement interactions have immediate implications on public health, which can only be realized by making the data easily accessible to any interested researcher, clinician or consumer.
We note that many medical providers in developing countries do not have subscriptions to clinical databases such as TRC\footnote{\href{https://naturalmedicines.therapeuticresearch.com/}{https://naturalmedicines.therapeuticresearch.com/}} and UpToDate,\footnote{\href{https://www.uptodate.com/}{https://www.uptodate.com/}} and may lack an easy way to identify possible supplement interactions before prescribing drugs to their patients. 
To fill this gap, we develop \suppai (available at \url{https://supp.ai/}), an application for browsing evidence of supplement interactions extracted from clinical and biomedical literature. \suppai allows users to: 
\begin{itemize}[itemsep=0.1mm]
    \item Search for supplements or drugs, 
    \item Search through potential interactions, 
    \item Browse evidence sentences with supplement and drug entities highlighted, 
    \item Navigate links to source papers 
\end{itemize}

We design \suppai to be a rapid way for users to access and search extracted SDI and SSI evidence. Our goal for this application is to provide a high quality, broadly-sourced, up-to-date, and easily accessible platform for searching through SDI and SSI evidence, while providing sufficient information for users to judge the quality of each piece of evidence. In Section \ref{sec:approach}, we describe the NLP pipeline used to extract evidence from scientific papers. Below, we describe the user interface and data features of \suppai.

\subsection{User interface}
\label{sec:ui_details}

Besides the main search page seen by users when they first navigate to the site, \suppai consists of two other types of pages: entity and interaction pages. Entity pages provide information about one supplement or drug, and a list of potential interacting entities, sorted by quantity of evidence. We provide information such as synonyms, drug trade names, and definitions about each entity upon hover over or expansion. Interaction pages display all discovered pieces of evidence supporting an interaction between a pair of entities. The evidence is sorted by additional features extracted from source papers, such as the level of evidence and recency, discussed in Section \ref{sec:supporting_data}. 

Figure \ref{fig:supp_ai} shows the interface, with results for the ginkgo supplement. Results on the entity page (\emph{left}) list 140 possible interactions to entities such as Warfarin and Nitric Oxide. When a result is selected, the interaction page is displayed (\emph{right}), showing evidence sentences supporting the interaction along with metadata and links to each source paper. Spans linked to supplement and drug entities in evidence sentences are highlighted. To see more context or detail about the interaction, the user can navigate to the source paper to continue reading.

\subsection{Supporting data for search}
\label{sec:supporting_data}

We extract additional paper metadata as a way to judge evidence quality. From Semantic Scholar, we retrieve the paper title, authors, publication venue, and year of publication. Medical Subject Headings (MeSH) tags associated with each paper are used to determine whether its results are derived from clinical trials, case reports, or animal studies. We also attempt to identify the retraction status of each paper, again using MeSH tags. Evidence sentences are ordered and presented based on associated paper metadata, prioritizing non-retracted studies, clinical trials, human studies, and recency (year of publication).

Using the RxNorm relationship \emph{has\_tradename} via the Unified Medical Language System (UMLS) Metathesaurus \citep{Bodenreider2004TheUM}, we derive trade names associated with drug ingredients, e.g. \emph{Prozac} and \emph{Sarafem} are trade names of the ingredient \emph{fluoxetine}. Trade drugs are associated with active drug ingredients and indexed for search. Users can query a trade name rather than an active ingredient and be directed to the relevant interactions.

\subsection{Data \& API}
\label{sec:data_api}

Data on the site are periodically updated as new papers are incorporated into the Semantic Scholar corpus. Snapshots of the data are available for download at \url{https://api.semanticscholar.org/supp/}. Live data on the site, which is updated more frequently, can be accessed through our search API, documented at \url{https://supp.ai/docs/api}. Additionally, we provide training data, evaluation data, and the curated drug/supplement identifier lists (discussed in Section \ref{sec:approach}) used to produce the dataset of interactions at \url{https://github.com/allenai/sdi-detection}. We encourage others to reuse our data and model to improve information availability around supplement interactions and safety.

\begin{figure*}[th!]
    \centering
    \includegraphics[width=\textwidth]{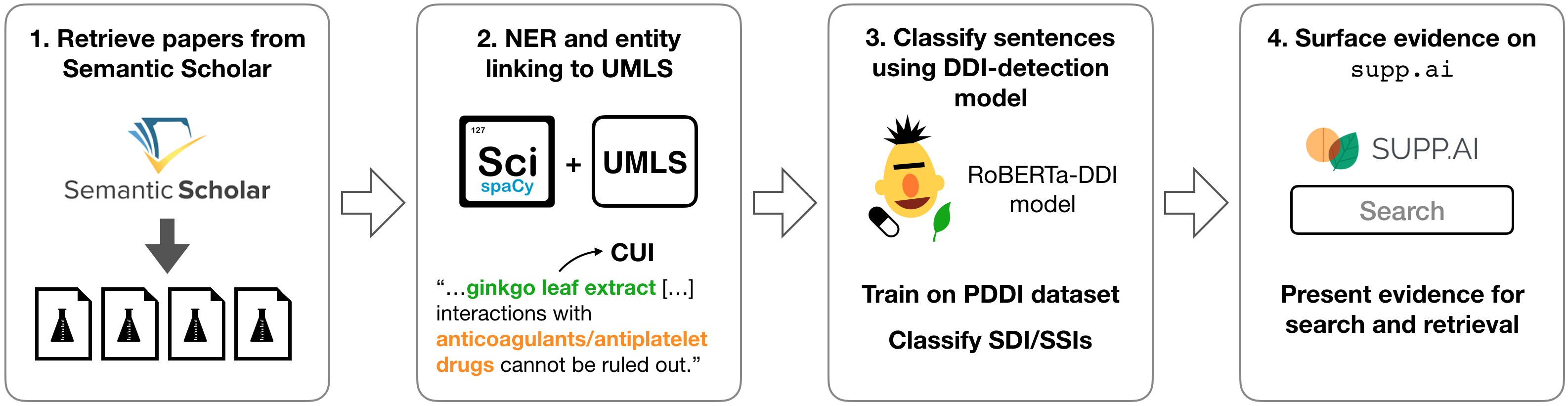}
    \caption{Pipeline for identifying sentences containing evidence of SDIs and SSIs.}
    \label{fig:pipeline}
\end{figure*}

\section{Methods}
\label{sec:approach}

An overview of our NLP pipeline is given in Figure \ref{fig:pipeline}. 
We first retrieve Medline-indexed articles using the Semantic Scholar API,\footnote{\href{https://api.semanticscholar.org/}{https://api.semanticscholar.org/}} and pre-process the text to generate candidate evidence sentences (Section \ref{sec:data_preprocessing}).
We then use our DDI-detection model, a neural network classifier based on BERT \citep{Devlin2018BERTPO} and fine-tuned on labeled DDI data from \citet{ayvaz_toward_2015} (Section \ref{sec:classifier}), to classify sentences for the existence of an interaction.
Sentences classified as positive by our model are collated and surfaced on \suppai (Section \ref{sec:supp_ai_browser}).

\subsection{Generating candidate evidence}
\label{sec:data_preprocessing}

Approximately 22M Medline-indexed articles are downloaded using the Semantic Scholar API. The scispaCy library \citep{Neumann2019ScispaCyFA} is used to perform sentence tokenization, NER, and entity linking over all paper abstracts. Entity mentions are linked to Concept Unique Identifiers (CUIs) from the UMLS Metathesaurus. An example sentence from \citet{Vaes2000InteractionsOW} is shown with linked entity mentions:

\begin{quote}
\small
\texttt{$\underbracket{\text{\textcolor{darkgray}{Hemorrhage}}}_{\text{C0019080}}$ and tendencies were noted in four $\underbracket{\text{\textcolor{darkgray}{cases}}}_{\text{C0868928}}$ with $\underbracket{\text{\textcolor{blue}{ginkgo}}}_{\text{C0330205}}$ use and in three $\underbracket{\text{\textcolor{darkgray}{cases}}}_{\text{C0868928}}$ with $\underbracket{\text{\textcolor{blue}{garlic}}}_{\text{C0017102}}$; in none of these $\underbracket{\text{\textcolor{darkgray}{cases}}}_{\text{C0868928}}$ were $\underbracket{\text{\textcolor{darkgray}{patients}}}_{\text{C0030705}}$ receiving $\underbracket{\text{\textcolor{blue}{warfarin}}}_{\text{C0043031}}$.}
\end{quote}

Of these linked entities, we preserve entities on a list of curated supplements and drugs (entities in blue). We generate these curated lists in a semi-automatic fashion, by querying the children of UMLS supplement and drug classes and performing fuzzy name matching to known supplements or drugs crawled from the web. We also perform clustering of similar entities to reduce redundancy in the final dataset, e.g., combining several variants of Vitamin D together into a single entity. Details on identifier curation and clustering are given in Appendix \ref{app:identifiers}.

We retain all sentences containing at least two entity mentions. For each sentence, we generate candidate evidence as each combination of two entity spans from that sentence.

\subsection{DDI-detection model}
\label{sec:classifier}

We train a DDI-detection model to predict whether a given candidate sentence provides evidence of an interaction between two drug entities. Our DDI-detection model uses pre-trained BERT models (Bidirectional Encoder Representations from Transformers) \citep{Devlin2018BERTPO} to encode input sequences. These models have been shown to be effective at domain transfer, and are able to achieve high performance using small amounts of task-specific annotated data. In particular, we use the large version of the pre-trained RoBERTa model, a further-optimized BERT model, that has approximately 340M parameters \citep{Liu2019RoBERTaAR}. We fine-tune the pre-trained embeddings of the RoBERTa language model using labeled data for DDI classification, and we call the resulting model RoBERTa-DDI.

\paragraph{Input layer:} The input layer consists of the sequence of byte-pair encoding word pieces \citep{Radford2019LanguageMA} in a sentence.
We replace entity mention spans with the special tokens \texttt{[Arg1]} and \texttt{[Arg2]}.
This helps generalization by preventing the model from memorizing entity pairs with positive interactions in the training set.
For example:
\begin{quote}
\small
\texttt{[CLS] Combination [Arg1] may also decrease the plasma concentration of [Arg2]. [SEP]}
\end{quote}

\noindent where \texttt{[Arg1]} and \texttt{[Arg2]} replace the spans ``hormonal contraceptives'' and ``acetaminophen'' respectively. We add special tokens \texttt{[CLS]} and \texttt{[SEP]} at the beginning and end of each sentence to leverage their representations learned in pre-training.
At prediction time, candidate sentences are masked similarly and fed to the trained model.

\paragraph{Model architecture:} As the name implies, RoBERTa-DDI uses the pre-trained RoBERTa representations \cite{Liu2019RoBERTaAR} to encode input sequences. We refer readers to \citet{Liu2019RoBERTaAR}, \citet{Devlin2018BERTPO}, and \citet{Vaswani2017AttentionIA} for more details on BERT and transformer architecture. For the RoBERTa-DDI model, we add a dropout layer followed by one feedforward (output) layer with a softmax non-linearity, which takes the representation of the \texttt{[CLS]} token at the top transformer layer as input and outputs probabilities for labels $\{0, 1\}$, where $1$ indicates an interaction.

\paragraph{Model training:}
Due to similarities between DDIs and SDIs/SSIs, we hypothesize that a classifier trained to identify DDI evidence should perform well in identifying SDI and SSI evidence. We therefore take advantage of existing labeled data for categorizing DDIs to fine-tune the model. We use pre-trained weights distributed by the authors of \citet{Liu2019RoBERTaAR}, and further fine-tune the model parameters (as well as parameters of the output layer) using labeled DDI data from the Merged-PDDI dataset \citep{ayvaz_toward_2015}. 

In particular, we use training data from the DDI-2013 \citep{SeguraBedmar2013SemEval2013T9} and NLM-DailyMed \citep{Stan2014TitleA} datasets, as they are relatively large and contain evidence sentences with annotated drug mention spans. The DDI-2013 dataset consists of sentences extracted from DrugBank and Medline; the NLM-DailyMed dataset draws sentences from cardiovascular drug product labels retrieved from DailyMed. Both datasets contain multi-class labels for different types of interactions. We distinguish between detection, a binary classification problem where the goal is to determine whether an interaction exists or not, and multi-class classification, where the goal is to determine the type of interaction. In this work, we focus on detection, but provide results for a variant of our model trained on classification that obtains SOTA performance compared to prior work. 

For detection, we collapse labels corresponding to all interaction types (e.g., mechanism, advise, effect, etc.) into binary labels of 0 and 1, where 0 means no interaction, and 1 means an interaction of some type exists. Collapsing the positive labels is necessary for training one DDI-detection model on both the DDI-2013 and NLM-DailyMed datasets, since the two datasets are annotated with inconsistent interaction types. We preserve the train/test splits used in \citet{ayvaz_toward_2015}, and create a development set from the training set for iteration on model design and tuning.

A sentence from the training data can contain multiple drug entities. For training, we generate pairwise combinations of drug mention spans in each sentence. We note that many sentences are seen multiple times by our model with different labeled spans. Due to combinatorial explosion, and to prevent our model from learning excessively from a few instances containing lots of entity mentions, we restrict the training data to sentences containing less than or equal to 100 pairwise entity combinations. Table \ref{tab:training_data} shows the resulting data splits for the two datasets.

\begin{table}[h]
    \centering
    \scalebox{0.9}{
    \begin{tabular}{lcccc}
        \toprule
        Dataset & Train & Dev. & Test & Label=1 \\
        \midrule
        DDI-2013 & 18362 & 2069 & 5688 & 17.2\% \\
        NLM-DailyMed & 11372 & 1255 & 927 & 22.7\% \\
        \bottomrule
    \end{tabular}
    }
    \caption{DDI training data split.}
    \label{tab:training_data}
\end{table}

Our training hyperparameters follow those presented by \citet{Liu2019RoBERTaAR} (learning rate = 1e-5; 4 epochs). No additional hyperparameter tuning is performed.

\section{Results \& evaluation}
\label{sec:results}
Of the 22M articles we retrieve, around 4.6M abstracts contain candidate sentences. After initial filtering, 33.0M candidate sentences containing supplement entity mentions are classified by RoBERTa-DDI. Around 625k (1.9\%) of these sentences are classified as positive for an interaction. We perform entity normalization across positive sentences based on CUI clusters, and perform additional ad hoc filtering of evidence to eliminate incorrectly detected spans resulting from poor NER and linking, such as the span ``retina'' linking to Vitamin A (C0040845). The resulting 195k sentences contain mentions of 2044 unique supplements and 2772 unique drugs, and provide evidence sentences for 60k interactions sourced from 133k papers.

Comparisons of model variants on DDI classification and detection (including SOTA results on both tasks) are given in Appendix \ref{app:sota_ddi}. To evaluate the transferability of DDI detection to the related task of SDI/SSI detection, we use a test set consisting of 500 sentences annotated for the presence or absence of a supplement interaction. To obtain a balanced test set despite the rare presence of a positive interaction, we sample half the instances from the set of sentences labeled as positive by a previous variant of our model based on fine-tuning BERT-large, and the other half from those labeled as negative. After manual annotation, 40\% of the sampled instances were positive for an interaction. Annotation was performed by two authors without seeing model predictions, with an inter-annotator agreement of 94\%. This test set was used for final evaluation, and never for model development or tuning. Table \ref{tab:eval_sdi} shows the performance of RoBERTa-DDI on the DDI and supplement test sets. Performance on the SDI test set has precision 0.82, recall 0.58, and F1-score 0.68. Although there is performance degradation during transfer, the precision of detection remains high at 0.82. 

Decrease in recall can be attributed to a larger percentage of positive instances in the SDI test set (roughly 40\%, compared to 20\% in the DDI training data). Another factor is the presence of incorrectly labeled entity spans in the supplements test set due to NER/linking errors. To better understand this second source of errors, we attempt to evaluate the performance of the scispaCy entity linker. Processing each sentence from the two DDI training sets using scispaCy, we determine that only 80\% of drug entities from DDI-2013 and 76\% from NLM-DailyMed are recognized and linked. 
The likelihood of supplement entities being successfully linked is likely lower, due to sparse training data for supplement NER and linking. 
These numbers provide an estimate of the global ceiling on recall for our model. In future work, we aim to explore ways to improve NER and linking and assess their impact on the results of SDI detection. SDI/SSI sentences in our output set can also be labeled by biomedical expert annotators and used to further tune the model for SDI/SSI detection.

\begin{table}[h]
    \centering
    \scalebox{0.95}{
    \begin{tabular}{lccc}
        \toprule
        Evaluation set & Prec. & Rec. & F1 \\
        \midrule
        Drugs (DDI-2013) & 0.90 & 0.87 & 0.88 \\
        Drugs (NLM-DailyMed) & 0.83 & 0.85 & 0.84 \\ 
        Supplements-500 & 0.82 & 0.58 & 0.68 \\
        \bottomrule
        \end{tabular}
        }
    \caption{The RoBERTa-DDI model (trained on drug-drug interaction labels) is evaluated on two DDI evaluation sets (first two rows) and our supplement interaction evaluation set (last row). }
    \label{tab:eval_sdi}
\end{table}

\section{Discussion}

Information describing the safety and efficacy of dietary supplements can be difficult to find. The inability to locate evidence of SDIs can challenge clinician ability to advise patients and cause risks for consumers of dietary supplements. It is our hope that extracting evidence for SDIs/SSIs from a large corpus of scientific literature and making the evidence available through an easily accessible search interface can offset some of these risks. 

This work demonstrates how NLP techniques can be extraordinarily useful for extracting information and relationships specific to an application domain in healthcare. Re-purposing existing labeled data from related domains (that would be expensive to generate in a new domain) can be a way to derive maximum utility from curation efforts. Continuing, we look to investigate fine-grained interaction types, and provide better classification of the level of evidence provided by each sentence or document towards a particular SDI or SSI. We also aim to leverage similar techniques for identifying evidence of indications, contraindications, and side effects of dietary supplements from the biomedical and clinical literature, and make these discoverable on \suppai.

\subsection{Related Work}

Consumer-facing websites such as the NIH Office of Dietary Supplements\footnote{\href{https://ods.od.nih.gov/}{https://ods.od.nih.gov/}} or WebMD\footnote{\href{https://www.webmd.com/vitamins/index}{https://www.webmd.com/vitamins/index}} provide facts about common supplements, but this information can be incomplete and may not support researcher or clinician needs. TRC Natural Medicines\footnote{\href{https://naturalmedicines.therapeuticresearch.com/}{https://naturalmedicines.therapeuticresearch.com/}} and UpToDate\footnote{\href{https://www.uptodate.com/}{https://www.uptodate.com/}}, two dedicated clinical resources, contain high-quality, curated evidence, but may not be broadly accessible due to their subscription format. Drug databases like DrugBank \cite{wishart_drugbank_2018}, RxNorm \cite{Nelson2011NormalizedNF}, and the National Drug File Reference Terminology (NDFRT) \cite{Simonaitis2010QueryingTN} contain only partial coverage of supplement terminology \cite{Manohar2015EvaluationOH}, and primarily focus on aggregating drug information.

Several prior studies have experimented with extracting safety information of supplements and supplement interactions from various forms of text. 
\citet{Zhang2015MiningBL} employ machine learning techniques to filter supplement interaction relationships in SemMedDB, a database of relationships extracted from Medline articles. 
\citet{Jiang2017DiscoveringPE} develop a model for identifying adverse effects related to dietary supplements as reported by consumers on Twitter, and discover 191 adverse effects pertaining to 4 dietary supplements. 
\citet{Fan2016ClassificationOU} and \citet{Fan2018UsingNL} analyze unstructured clinical notes to predict whether a patient started, continued or discontinued a dietary supplement, which can be useful as a building block for identifying adverse effects in clinical notes (as attempted by the same authors in \citet{Fan2017DetectingSO} for the drug warfarin).
\citet{Wang2017MiningAE} proposes using topic models to analyze the adverse effects of dietary supplements as mentioned in the Dietary Supplement Label Database, and finds that Latent Dirichlet Allocation models \cite{Blei2003LatentDA} can be used to group dietary supplements with similar adverse effects based on their labels. As far as we know, there are no other studies investigating the task of sentence-level identification of SDI/SSI evidence from the scientific literature. No previous work has investigated the utility of using labeled DDI data for transfer learning to SDI/SSI identification.

\subsection{Limitations}

There are several limitations of this work. First, we distinguish between supplements and drugs. Both supplements and drugs are pharmacologic entities, with their separate classification more attributable to marketing and social pressures rather than functional differences. However, due to this somewhat arbitrary distinction, supplement entities are not well represented in databases of pharmaceutical entities, and less information is publicly available on their interactions. We also use UMLS CUIs as a way of identifying supplement and drug entities. The lack of a standardized terminology to describe dietary supplements is discussed in \citet{manohar_evaluation_2015} and \citet{wang_term_2016}, which estimate UMLS coverage of these terms to be between 14-54\%. This limitation prevents us from identifying many supplement entities. Lastly, our dependence on NLP-pipeline tools sets a performance ceiling due to unsolved problems in NER and linking. Although scispaCy is performant and detects a large number of relevant entities, our evaluations show that many supplement and drug entities are missed. A system such as MetaMapLite \citep{DemnerFushman2017MetaMapLA} has higher recall, but performance is slow and there are practical challenges to using it to process large numbers of documents.

\subsection*{Conclusion}

Insufficient regulation in the supplement space introduces dangers for the many users of these supplements. Claims of interactions are difficult to validate without links to source evidence. We create an NLP pipeline to detect SDI/SSI evidence from scientific literature, leveraging UMLS identifiers, scispaCy for NER and entity linking, BERT-based language models for classification, and labeled data from a related domain for training. We use this pipeline to extract evidence from 22M biomedical and clinical articles with high precision. The extracted SDI/SSI evidence are made search-able through a public web interface, \suppai, where we integrate additional metadata about source papers to help users make decisions about the reliability of evidence. Our dataset and web interface can be leveraged by researchers, clinicians, and curious individuals to increase understanding about supplement interactions. We hope to encourage additional research to improve the safety and benefits of dietary supplements for their consumers.

\subsubsection*{Acknowledgments}
We would like to thank Oren Etzioni for his indispensable feedback and support of this project. We thank Amandalynne Paullada for contributing to an earlier prototype, and we thank Asma Ben Abacha, Pieter Cohen, Taha Kass-Hout, Beth Ranker, Lia Schmitz, Heidi Tafjord, and our users for helpful comments on improving \suppai.

\bibliography{whatsupp-2019}
\bibliographystyle{acl_natbib}

\appendix

\begin{table}[tbph!]
    \centering
    \scalebox{0.75}{
    \begin{tabular}{p{35mm}p{15mm}p{18mm}p{18mm}}
        \toprule
        Test dataset & Num. pairwise instances & RoBERTa-DDI (Trained on DDI-2013 and NLM-DailyMed) & RoBERTa-DDI (Trained on DDI-2013 only) \\
        \midrule
        DDI-2013 (All) & 5688 & 0.88 & 0.89 \\
        DDI-2013 (DrugBank) & 5251 & 0.89 & 0.90 \\
        DDI-2013 (Medline) & 437 & 0.73 & 0.77 \\
        NLM-DailyMed & 927 & 0.84 & 0.70 \\
        \midrule
        All & 6615 & 0.87 & 0.85 \\
        \bottomrule
    \end{tabular}
    }
    \caption{F1-scores of RoBERTa-DDI trained using different training data. Test data contains all pairwise combinations of entities in test sentences.}
    \label{tab:ddi_eval}
\end{table}

\begin{table*}[tb!]
    \centering
    \small
    \begin{tabular}{p{69mm}p{30mm}p{25mm}p{12mm}}
        \toprule
        Model & Reference & F1 (classification) & F1 (detection) \\
        \midrule
        Bi-LSTM (w/ max and attentive pooling) & \citet{Sahu2017DrugDrugIE} & 0.69 (macro-F1) & - \\
        Hierarchical Bi-LSTM + Attention + dependency path & \citet{Zhang2018DrugdrugIE} & 0.73 (unspecified) & - \\
        Bi-LSTM (w/ attention and negative instance filtering) & \citet{Zheng2017AnAE} & 0.77 (unspecified) & 0.84 \\
        BioBERT embeddings & \citet{Chauhan2019REflexFF} & 0.72 (macro-F1) & 0.87 \\
        BERT-large embeddings fine-tuned on DDI-2013 & \citet{Peng2019TransferLI} & 0.79 (micro-F1) & - \\
        RoBERTa-DDI fine-tuned on DDI-2013 (Ours) & - & 0.82 (micro-F1) & 0.89 \\
        \bottomrule
    \end{tabular}
    \caption{Baseline models for DDI detection and reported performance on the DDI-2013 test set. Results are shown for classification (5-way classification) and detection (binary classification).}
    \label{tab:ddi_sota}
\end{table*}

\section{Supplement and drug identifiers}
\label{app:identifiers}

We generate lists of supplement and drug entities based on UMLS Concept Unique Identifiers (CUIs) using a semi-automated method. For supplements, we identify NCI thesaurus (NCIT) concepts such as ``Dietary Supplement'' (NCIT: C1505, CUI: C0242295), ``Vascular Plant'' (NCIT: C14336, CUI: C0682475), and ``Antioxidant'' (NCIT: C275, CUI: C0003402) as likely parents of supplement terms. We recursively extract child entities of these parent classes from UMLS, deriving an initial list of supplements. To improve recall, we extract supplement names from the TRC Natural Medicines database,\footnote{\href{https://naturalmedicines.therapeuticresearch.com/}{https://naturalmedicines.therapeuticresearch.com/}} perform fuzzy string matching to entities in UMLS, and add any identified CUIs to our list of supplements. The list is manually reviewed to remove non-supplement entities, those for which we could not identify any marketed supplement or medicinal uses. Following curation, we retain 2139 unique supplement entities.

Similarly, we generate a corresponding list of drug CUIs from parent entity ``Pharmacologic Substance'' (NCIT: C1909, CUI: C1254351) and any UMLS entity with a DrugBank identifier. Fuzzy name matching between drugs on drugs.com\footnote{\href{https://drugs.com/}{https://drugs.com/}} and UMLS entities is used to identify drugs and experimental chemicals missed through UMLS search alone. Due to the significantly larger number of drugs compared to supplements, manual curation of this list is impractical at this time. This process generates a list of 15252 unique drug CUIs. Any entity that is identified as both a supplement and a drug is categorized exclusively as a supplement for the purposes of this work.

Similar supplement and drug entities are merged, such as those with overlapping names, e.g., entities corresponding to UMLS C0006675, C0006726, C0596235, and C3540037 all describe variants of Calcium and are merged under the supplement entity C3540037 (“Calcium Supplement”). The canonical CUI representing a cluster is selected manually. Drug, supplement, and canonical mappings are provided in our data repository.

\section{DDI model performance}
\label{app:sota_ddi}

We train RoBERTa-DDI on a combination of DDI-2013 and NLM-DailyMed training data. In Table \ref{tab:ddi_eval}, we report the F1-scores of model variants on the test data. We show the performance of the final variant of RoBERTa-DDI (trained on both DDI-2013 and NLM-DailyMed) as well as a variant trained only on DDI-2013 training data (last column), which performs best on the DDI-2013 test set, but suffers when tested on NLM-DailyMed. We also further break down performance on the DrugBank and Medline sub-corpora within DDI-2013. 

The DDI-2013 dataset is used as a benchmark dataset for DDI detection and classification, and is part of the BLUE benchmark suite \citep{Peng2019TransferLI}. RoBERTa-DDI outperforms recently-reported SOTA performance on DDI detection in the DDI-2013 dataset using BioBERT \citep{Lee2019BioBERTAP} (F1 = 0.87) \citep{Chauhan2019REflexFF}. \citet{Peng2019TransferLI} also report SOTA performance on the DDI-2013 classification task, achieving 0.79 micro-F1 using a tuned BERT-large model. For comparison, we show the results of RoBERTa-DDI trained on DDI-2013 multi-class classification, which achieves 0.82 micro-F1 on DDI-2013 classification. We provide previously reported SOTA performance metrics on DDI-2013 in Table \ref{tab:ddi_sota}. We note that because the interaction classes are unbalanced in the DDI-2013 dataset, reported classification micro- and macro-F1-scores in previous work are not directly comparable.

The inclusion of the NLM-DailyMed corpus increases training data diversity and should improve generalization for the task of detecting SDI/SSI evidence. Thus, although RoBERTa-DDI trained on DDI-2013 has the highest performance on the DDI-2013 test set, RoBERTa-DDI trained over all training data performs the best overall, and we use this model variant to classify evidence for \suppai.

\end{document}